\documentclass{IEEEtran}

\usepackage{microtype}
\usepackage{graphicx}
\usepackage{booktabs}
\usepackage{mathtools}
\usepackage{amsmath,amssymb,amsfonts}
\usepackage{subfig}
\usepackage{multirow}
\usepackage{algorithm}
\usepackage{algorithmic}
\usepackage{hyperref}

\title{Privacy-Preserving Self-Taught Federated Learning for Heterogeneous Data}

\author{Kai-Fung Chu, Lintao Zhang\\
BaseBit Technologies\\
kaifungchu@gmail.com, lintao.zhang@basebit.ai}

\begin{document}
\maketitle

\begin{abstract}
Many application scenarios call for training a machine learning model among multiple participants. Federated learning (FL) was proposed to enable joint training of a deep learning model using the local data in each party without revealing the data to others. Among various types of FL methods, vertical FL is a category to handle data sources with the same ID space and different feature spaces. However, existing vertical FL methods suffer from limitations such as restrictive neural network structure, slow training speed, and often lack the ability to take advantage of data with unmatched IDs. In this work, we propose an FL method called self-taught federated learning to address the aforementioned issues, which uses unsupervised feature extraction techniques for distributed supervised deep learning tasks. In this method, only latent variables are transmitted to other parties for model training, while privacy is preserved by storing the data and parameters of activations, weights, and biases locally. Extensive experiments are performed to evaluate and demonstrate the validity and efficiency of the proposed method.
\end{abstract}

\section{Introduction}\label{intro}

Due to the advancement of artificial intelligence (AI) and deep learning \cite{DL_Nature} technologies, many human-level intelligent applications empowered by AI have emerged, such as object recognition \cite{ImageNet}, playing video games \cite{HumanLevelControl}, and playing Go \cite{AlphaGo}. To obtain a deep learning model with superior performance, a big dataset is one of the prerequisites in many applications. However, obtaining such a big dataset is a difficult task. The situation gets even worse for supervised tasks using labeled data since it may involve expensive human effort for labeling the data. It is very hard for a single party to gather enough training data. To avoid manually collecting all data by just one party, one may form a big dataset by combining the data and resources from distributed parties. Take health care as an example, data such as symptoms, medical records, and patient profiles can contribute to the AI medical diagnosis. However, different person may seek a doctor at different hospitals. Each hospital tends to keep the data locally, and thus the complete medical records are scattered. They may refuse to publish or share the data with others since personal and sensitive information may be revealed from the data and forbidden by rules and regulations. As a result, although combining the dataset may enhance the AI model, the data are still dispersed in different parties and forms data silos. Therefore, the isolated data problem and data privacy issues have to be carefully addressed for improving the AI model.

Many privacy-preserving computing technologies have been proposed to address the privacy issues in data sharing, namely, Differential Privacy, Homomorphic Encryption, and Secure Multi-Party Computation.
Differential Privacy \cite{dwork2008differential} enables sharing general information of the dataset without leaking individuals information in the dataset. However, there is a trade-off between the security level and accuracy.
Homomorphic Encryption \cite{gentry2009fully} is an encryption technique that allows computation to be performed over encrypted data without decryption. The decrypted result is equal to the results that are computed without any encryption. Thus, the encrypted data can be combined to form a large dataset without information leakage.
Secure Multi-Party Computation \cite{yao1986generate} is a method to enable jointly compute a function with input from different parties while the inputs are kept private. 
However, the above technologies cannot entirely support privacy-preserving joint training of a neural network-based model with distributed data sources due to their high computational cost and limited tool chain support.

To address the data privacy issues in machine learning, Google introduced a privacy-preserving machine learning framework called federated learning (FL) \cite{mcmahan2017communication} \cite{konevcny2016federated}, allowing different parties to jointly train a model without data sharing. Each party retains a copy of the model and trains the model using their local dataset. For each training iteration, only gradients are transmitted to other parties so that the parties cannot acquire the raw data from these parameters. This method is effective for parties with homogeneous data, i.e., the data with the same feature space. However, the feature space of the data stored in different parties may not always be the same. For example, a hospital may have a patient's medical record while the smart wearing device keeps track of the heartbeat rate and blood pressure. The data in different parties can be heterogeneous such that sharing of model parameters using the aforementioned method is not applicable in these cases. Developing an efficient FL method for heterogeneous data remains challenging. 

In \cite{zhang2018gelu} and \cite{zhang2019additively}, homomorphic encryption-based FL approaches are introduced to handle the heterogeneous data. In these approaches, a deep neural network is partitioned into two parts: feature extractor and classifier. Each guest party feeds their data to the feature extractor locally, and the encrypted outputs are provided to the classifier at the host party. The error and gradients are then computed based on the label data at the host. In the backward propagation stage, the encrypted gradients are sent back to the guests for their respective parameter updates. The intermediate results exchanged in different parties are homomorphicly encrypted or noise masked such that other parties cannot infer the raw data. However, these approaches suffer from some drawbacks. First, the design of the neural network structure is restricted due to the partitioning of the feature extractor and classifier in the guest and host, respectively. Second, the processing speed of these approaches is quite low due to the intensive homomorphic encryption and decryption among the parties. Finally, the data with unmatched IDs are discarded and wasted. Therefore, it is a promising research direction to develop a new FL approach that addresses the limitations for privacy-preserving multi-party machine learning problems with heterogeneous data.

In this work, inspired by self-taught learning \cite{raina2007self}, we propose a FL method called self-taught federated learning (STFL), which uses unsupervised feature extraction techniques for distributed supervised deep learning tasks. In this method, only latent variables are transmitted to other parties for model training while privacy is preserved by storing the data and parameters of activations, weights and biases locally. Extensive experiments are performed to evaluate and demonstrate the validity and efficiency of the proposed method.

\section{Related Work}
In this section, the related works of FL and self-taught learning are reviewed.

\subsection{Federated Learning}
FL is a privacy-preserving machine learning framework designed by Google \cite{mcmahan2017communication} \cite{konevcny2016federated} to address this problem. Depending on the data structures, the FL methods can be divided into three categories as defined in \cite{yang2019federated}: horizontal federated learning, vertical federated learning, and federated transfer learning.

\subsubsection{Horizontal Federated Learning}
\begin{figure}[!t]
\centering
\includegraphics[width=0.48\textwidth]{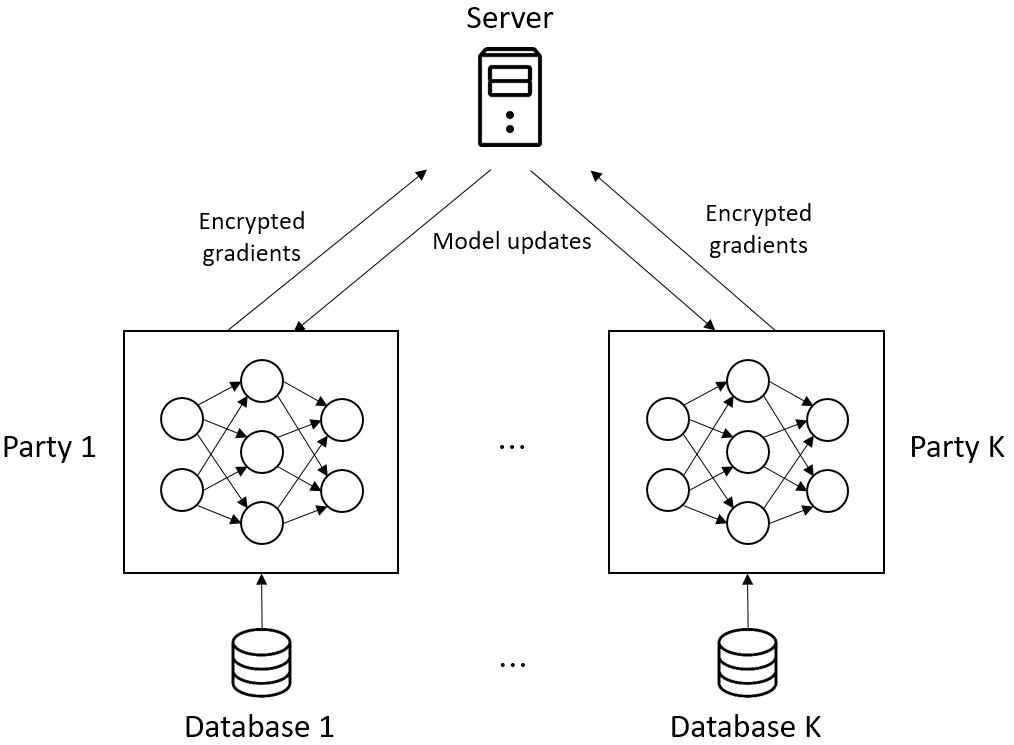}
\caption{Horizontal federated learning.}\label{horizontal_fig}
\end{figure}

Consider the data is homogeneous, implying the feature spaces of data in different parties are the same. The learning task in this scenario can be categorized as horizontal federated learning. For example, two regional hospitals may have patient medical records in the same feature space. The group of the patient may be diverse so that the intersection of the dataset is small. In this case, combining the dataset is generally beneficial to the AI model training by increasing the number of samples. In \cite{mcmahan2017communication}, each party trains a local deep learning model using their local dataset and uploads the trained parameters to the server as shown in Figure \ref{horizontal_fig}. The server updates the global model by calculating the average of the parameters and broadcasts the updated global model to the parties in each iteration. Using this method, the global model is equivalent to being trained by all the data in the parties, while the data itself is stored locally without sharing to other parties. The only parameters sharing with the server are the gradients. The parties cannot infer the raw data based on gradients. Hence, the data privacy is preserved in this algorithm.

\subsubsection{Vertical Federated Learning}
Vertical federated learning is usually used to deal with heterogeneous data, which refers to the machine learning tasks where the data in the parties contain the same IDs, but different feature spaces. In \cite{zhang2018gelu}, a privacy-preserved machine learning framework based on homomorphic encryption is introduced. Based on the findings in \cite{zhang2018gelu}, the authors in \cite{zhang2019additively} extend the capability of the framework to support the scenario where the data and labels are distributed in different parties. The training of deep neural network carries a partly unencrypted and partly encrypted strategy between different parties to avoid learning on encrypted data directly. There are two stages in the training: forward propagation and backward propagation as shown in Algorithm \ref{VFL_forward} and \ref{VFL_backward}, respectively. The former is responsible for forwarding the encrypted activations from the guests to the host, while the later is responsible for propagating the noise masked gradient to the guests for model update. The reason of encrypting or noise masking the intermediate information is that one may infer other's data by stacking the information of activations, model weights, and gradients. Since the parties cannot get the unencrypted intermediate information, they cannot infer other's data during the training process by using this approach. Hence, the data privacy issue is addressed by this approach.

\begin{algorithm}[tb]
\caption{Vertical FL Forward Propagation}
\label{VFL_forward}
\textbf{Input}: data $x$\\
\textbf{Initialize} accumulated noise $\epsilon_{acc}$, noise masked host's weight $\widetilde{W}_h$, guest's weight $W_g$
\begin{algorithmic}
\STATE \textbf{Guests:}
\STATE $a_i \leftarrow \sigma(W_g x)$
\STATE $[a_i]_c \leftarrow Enc(a_i)$
\STATE Send $[a_i]_c$ to host
\STATE \textbf{Host:}
\STATE $[\widetilde{z_i}]_c \leftarrow \widetilde{W}_h \otimes [a_i]_c$
\STATE $[\widetilde{z_i} + \epsilon_s]_c \leftarrow [z_i]_c \oplus \epsilon_s$
\STATE Send $[\widetilde{z_i} + \epsilon_s]_c$ to guests
\STATE \textbf{Guests:}
\STATE $\widetilde{z_i} + \epsilon_s \leftarrow Dec([\widetilde{z_i} + \epsilon_s]_c)$
\STATE $z_i + \epsilon_s \leftarrow (\widetilde{z_i} + \epsilon_s) - \epsilon_{acc} a_i$
\STATE Send $z_i + \epsilon_s$ to host
\STATE \textbf{Host:}
\STATE $z_i \leftarrow (z_i + \epsilon_s) - \epsilon_s$
\STATE $\hat{y}_i$ = Softmax$(z_i)$
\end{algorithmic}
\end{algorithm}

\begin{algorithm}[tb]
\caption{Vertical FL Backward Propagation}
\label{VFL_backward}
\textbf{Input}: Prediction $\hat{y}_i$, Label $y_i$, encrypted activation $[a_i]_c$, learning rate $\eta$
\begin{algorithmic}
\STATE \textbf{Host:}
\STATE $\frac{\partial L_i}{\partial \hat{y}_i} \leftarrow y_i - \hat{y}_i$
\STATE $[\frac{\partial L_i}{\partial W_h}]_c \leftarrow \frac{\partial L_i}{\partial \hat{y}_i} \otimes [a_i]_c$
\STATE $\frac{\widetilde{\partial L_i}}{\partial a_i} \leftarrow \frac{\partial L_i}{\partial \hat{y}_i} \widetilde{W}_h$
\STATE $[\frac{\partial L_i}{\partial W_h} + \epsilon_s]_c \leftarrow [\frac{\partial L_i}{\partial W_h}]_c \oplus \epsilon_s$
\STATE Send $[\frac{\partial L_i}{\partial W_h} + \epsilon_s]_c$ to guests
\STATE \textbf{Guests:}
\STATE $\frac{\partial L_i}{\partial W_h} + \epsilon_s \leftarrow Dec([\frac{\partial L_i}{\partial W_h} + \epsilon_s]_c)$
\STATE $\frac{\widetilde{\partial L_i}}{\partial W_h} + \epsilon_s \leftarrow (\frac{\partial L_i}{\partial W_h} + \epsilon_s) - \frac{\epsilon_w}{\eta}$
\STATE $[\epsilon_{acc}]_c \leftarrow Enc(\epsilon_{acc})$
\STATE $\epsilon_{acc} \leftarrow \epsilon_{acc} + \epsilon_w$
\STATE Send $\frac{\widetilde{\partial L_i}}{\partial W_h} + \epsilon_s$ and $[\epsilon_{acc}]_c$ to host
\STATE \textbf{Host:}
\STATE $\frac{\widetilde{\partial L_i}}{\partial W_h} \leftarrow (\frac{\widetilde{\partial L_i}}{\partial W_h} + \epsilon_s) - \epsilon_s$
\STATE $\widetilde{W}_h \leftarrow \widetilde{W}_h - \eta \frac{\widetilde{\partial L_i}}{\partial W_h}$
\STATE $[\frac{\partial L_i}{\partial a_i}]_c \leftarrow \frac{\widetilde{\partial L_i}}{\partial a_i} - [\epsilon_{acc}]_c \otimes \frac{\partial L_i}{\partial \hat{y}_i}$
\STATE Send $[\frac{\partial L_i}{\partial a_i}]_c$ to guests
\STATE \textbf{Guests:}
\STATE $\frac{\partial L_i}{\partial a_i} \leftarrow Dec([\frac{\partial L_i}{\partial a_i}]_c)$
\STATE Perform backpropagation with $\frac{\partial L_i}{\partial a_i}$
\end{algorithmic}
\end{algorithm}

\subsubsection{Federated Transfer Learning}
Federated transfer learning is designed for the scenario where both the ID and feature space are different. The joint training without overlapping in either ID or feature space is different from the above-mentioned algorithms, and thus techniques in transfer learning are usually applied. In \cite{gao2019privacy}, a secure feature mapping is used to transform the uncommon feature spaces of the parties into common ones before applying horizontal federated learning to the homogenized data. The authors in \cite{li2019fedmd} developed a federated transfer learning framework via model distillation to allow the parties to have their unique neural network model. In \cite{liu2020secure}, the authors incorporate additively homomorphic encryption and secret sharing into the federated transfer learning framework for transferring knowledge from source to target domain. In \cite{peng2020federated}, a model called federated adversarial domain adaptation is proposed to transfer the learned knowledge in the domains of the parties to an unlabeled target domain while persevering the data privacy in each party. The target model parameters are updated from the aggregation of the gradients of the separately trained local model using a dynamic attention mechanism.

\subsection{Self-Taught Learning}
Self-taught learning \cite{raina2007self} is a machine learning framework using unlabeled data in supervised classification tasks to improve classification performance. The main concept of self-taught learning is to use a large amount of unlabeled data of other object classes randomly collected from other data sources to construct high-level feature representations using sparse coding. The self-taught learning algorithm involves solving the following optimization problem:
\begin{equation}\label{STL_Eq}
\begin{split}
    \min_{b,a} & \quad \sum_i ||x^{(i)}_u - \sum_j a^{(i)}_j b_j ||^2_2 + \beta ||a^{(i)}||_1 \\
    \text{s.t.} & \quad ||b_j||_2 \leq 1, \quad \forall j \in 1, \ldots, s
\end{split}
\end{equation}
where $\{x^{(1)}_u, \ldots, x^{(k)}_u\}$ is the unlabeled data with each $x^{(1)}_u \in \mathbb{R}^n$, $b=\{b_1, b_2, \ldots, b_s\}$ is the basis vectors with each $b_j \in \mathbb{R}^n$, $a=\{a^{(i)}\}$ is the activations with each $a^{(i)} \in \mathbb{R}^s$, $n$ is the input dimension, and $s$ is the number of bases. Eq. (\ref{STL_Eq}) aims to minimize the difference between the input $x^{(i)}_u$ and linear combination of the bases $b_j$ and activations $a^{(i)}_j$ which encourages to reconstruct the input, and to minimize $a^{(i)}_j$ to make it sparse.

After constructing the set of bases $b$ using sparse coding, it computes $\hat{a}(x^{(i)}_l) \in \mathbb{R}^s$ for every input $x^{(i)}_l \in \mathbb{R}^n$ based on Eq. (\ref{STL_argmin}):
\begin{equation}\label{STL_argmin}
    \hat{a}(x^{(i)}_l) = \mathop{\arg\min}_{a^{(i)}} ||x^{(i)}_l - \sum_j a^{(i)}_j b_j||^2_2 + \beta||a^{(i)}||_1
\end{equation}
The equation approximates the input $x^{(i)}_l$ using a sparse linear combination of the bases $b_j$. The sparse vector $\hat{a}(x^{(i)}_l)$ becomes the new expression of the inputs $x^{(i)}_l$ which uses as inputs for the supervised classification task.

\section{Problem Definition}\label{problem_definition}
We define the isolated data problem in this section. Considering a set of guest party $\mathcal{K}$, each party $k \in \mathcal{K}$ has a private dataset $\mathcal{D}_k = \{x^{(i)}_k, \ldots, x^{(i+N_k-1)}_k\}$ where $x^{(i)}_k$ is the data in feature space $\mathcal{X}_k$ and $N_k$ is the number of data samples in party $k$. The superscript $(i)$ of the data represents a unique ID in the ID space $\mathcal{I}_k$. In particular, a party $H$ who contains the data in target label space $\mathcal{Y}$ acts as the host. For notational simplicity, we represent the dataset of the host as $\mathcal{D}_H = \{x^{(1)}_H, y^{(1)}, \ldots, x^{(N_H)}_H, y^{(N_H)}\}$ where $y^{(i)}$ is the label data and $N_H$ is the number of data samples in the host.

For vertical FL, each guest party must have some ID overlapped with the ID in the host to classify as heterogeneous data. In other words, the intersection of ID space between host and any guest party is not null:
\begin{equation}\label{IDSpace}
    \mathcal{I}_k \cap \mathcal{I}_H \neq \emptyset, \quad \forall k \in \mathcal{K}.
\end{equation}
To ensure each guest party contributing data in new feature space compared to the host, we consider the feature space of each $k$ subtract that of the host is not null:
\begin{equation}\label{FeatureSpace}
    \mathcal{X}_k \setminus \mathcal{X}_H \neq \emptyset, \quad \forall k \in \mathcal{K}.
\end{equation}
The party with the dataset satisfied Eqs. (\ref{IDSpace}) and (\ref{FeatureSpace}) can contribute new data samples during the joint model training where the datasets are defined as heterogeneous data.

All participating parties aim to jointly train a neural network-based machine learning model for a classification task on the label of the host while the privacy of its own dataset are maintained. The prediction accuracy of the jointly trained model should be similar to that of the model centrally trained by all dataset $\mathcal{D}=\{\mathcal{D}_1, \ldots, \mathcal{D}_{|K|}, \mathcal{D}_H\}$.

We assume the parties are semi-honest, which is a commonly used adversary model in privacy-preserving computing \cite{brickell2005privacy}. In this model, each party merely follows the protocol and will not deviate from the protocol to cheat. However, they are curious about the data input of others and can use any obtained intermediate information to infer the data. Hence, any unpremeditated information leakage can be prevented by achieving this level of security.

\section{Methodology}
In this section, we introduce the proposed methodology for accomplishing the vertical FL task with heterogeneous data scenario as stated in Section \ref{problem_definition} and present the security analysis of the proposed method.

\subsection{Self-Taught Federated Learning}
To address the issues discussed in Section \ref{intro}, we aim to achieve the following goals in this work. Firstly, a flexible vertical federated learning framework will be developed for admitting any neural network models and training algorithms. Secondly, the framework will be homomorphic encryption free to enhance the processing speed while preserving data privacy. Finally, the framework should leverage non-overlapped and public data of individual parties during the training process.

Inspired by self-taught learning, we propose a vertical FL framework called self-taught federated learning (STFL), which uses unsupervised feature extraction for distributed supervised machine learning tasks. Self-taught learning \cite{raina2007self} is a machine learning framework for supervised tasks that use random unlabeled data to enhance performance. The motivation of self-taught learning is that it is hard to collect a large number of labeled images for the supervised classification tasks. Therefore, it uses unlabeled random images to pre-train the model and obtains high-level features based on sparse coding \cite{olshausen1996emergence}. Then, the pre-trained activation functions are transferred to the model to compute the features for the classification task. Hence, the classification performance is significantly improved with the same amount of labeled data. We employ the same concept for high-level feature extraction using additional unlabeled data for vertical FL.

\begin{figure}[!t]
\centering
\includegraphics[width=0.48\textwidth]{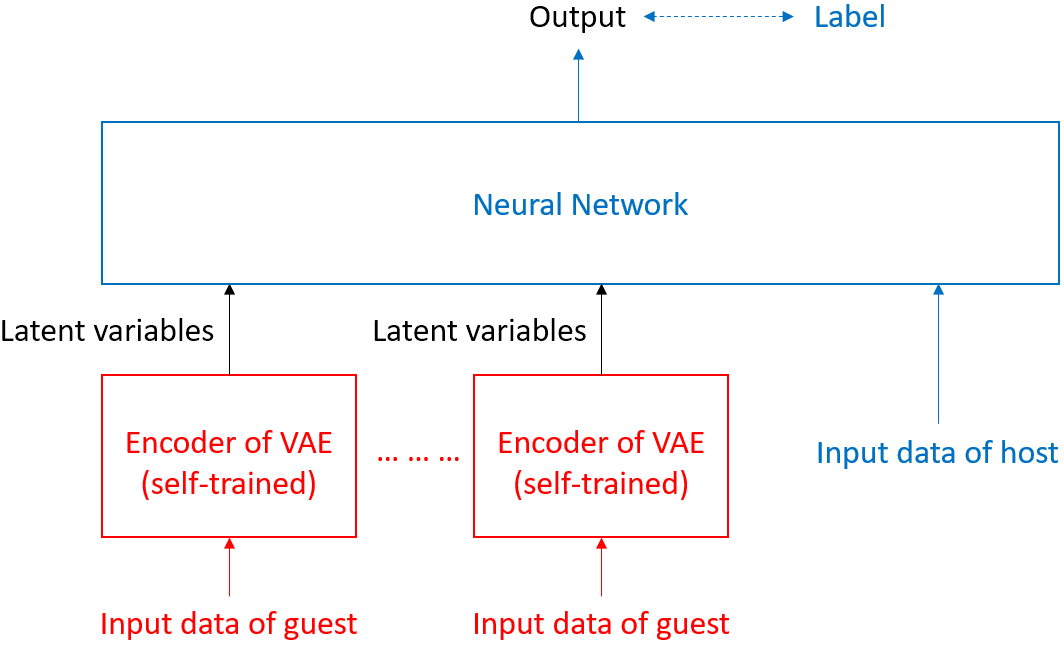}
\caption{Architecture of self-taught federated learning.}\label{architecture_fig}
\end{figure}

Figure \ref{architecture_fig} shows the architecture of STFL. It contains a neural network and encoders. The input data of host, neural network and label (in blue) are known to the host only while input data of guest and encoder (in red) are known to the corresponding guest only. Latent variables are the information shared from guest.
The detailed algorithm is shown in Algorithm \ref{algorithm}.
To preserve data privacy in the training process, each guest first trains a local encoder using their non-overlapped or public data based on feature extraction techniques such as variational autoencoder (VAE) \cite{kingma2014autoencoding}. We replace the high-level feature extraction technique from sparse coding in self-taught learning to VAE, which improves the security level. This reason will be provided in Section \ref{security_analysis}. The guests then output the latent variables from the encoder of VAE instead of directly output the data to other parties for model training.
More specifically, the VAE has the following loss function to minimize the KL-divergence and the reconstruction error:
\begin{equation}\label{VAE}
L(\phi, \psi) = D_{KL}(q_\phi(z|x)||p_\psi(z)) - E_{q_\phi(z|x)}[\log p_\psi(x|z)]
\end{equation}
where $q_\phi(z|x)$ is the encoder function for the posterior distribution and $p_\psi(x|z)$ is the decoder function for the likelihood distribution. The encoder of the guests may use a trainable activation function \cite{chung2016deep} to enhance the computational security further.

After training of the VAEs in the guests, the encoder can be used to encode the data to the latent space for joint training. The host and guests perform intersection protocols \cite{liang2004privacy} to compute the intersecting ID set among the parties without leakage of the set information. Only the data with the intersected ID will be used in the joint training. For the host, it is responsible for designing the master model, such as the neural network structure based on the data structure of the parties. It trains the model using the guests' input of latent variables and its own data with the same ID in each iteration. The training algorithm can be designed and implemented by the host while the guests only transmit latent variables from the encoder. The rest of the training is the same as common supervised learning. This algorithm preserves data privacy in each party as the raw data is not shared with others. The raw data cannot be inferred from the shared intermediate information. Detailed security analysis will be introduced in Section \ref{security_analysis}.

\begin{algorithm}[tb]
\caption{Self-Taught Federated Learning}
\label{algorithm}
\textbf{Input}: Unlabeled data of guests $k$ $\mathcal{D}_k$,\\
Labeled data of host $H$ $\mathcal{D}_H$\\
\textbf{Parameters}: VAE $\phi$, $\psi$ and master model $\theta$\\
\textbf{Output}: Trained parameters $\theta$\\
\textbf{Initialization}
\begin{algorithmic}
\STATE \textbf{Guests:}
\FOR{$k = 1$ to $|\mathcal{K}|$}
\STATE Train VAE $q_\phi$ and $p_\psi$ using $\mathcal{D}_k, \forall i \notin \{1, \ldots, N_H\}$ according to Eq. (\ref{VAE})
\ENDFOR
\STATE \textbf{Joint training}:
\FOR{$i=1$ to $N_H$}
\STATE \textbf{Guests:}
\FOR{$k = 1$ to $|\mathcal{K}|$}
\STATE $a^{(i)}_k \leftarrow q_\phi(x^{(i)}_k)$
\STATE Send $a^{(i)}_k$ to host
\ENDFOR
\STATE \textbf{Host:}
\STATE $\hat{y}^{(i)} \leftarrow f_{\theta}([x^{(i)}_H, a^{(i)}_k]))$
\STATE Update $\theta$
\ENDFOR
\STATE \textbf{return} $\theta$
\end{algorithmic}
\end{algorithm}

\subsection{Security Analysis}\label{security_analysis}
The computational and statistical security of the proposed algorithm is analyzed in this section.

\subsubsection{Computational Security}
Recall that each party aims to prevent the others from inferring their data. Although the data is not shared with others, information leakage may occur if the intermediate information are mishandled. During the training, the information shared is the latent variables from guest parties and final outputs from the host. For the guest, the obtained final outputs are calculated in the host by
\begin{equation}\label{host_forward}
    \hat{y} = \phi_H (W_H ( a^{(i)}_1, \ldots, a^{(i)}_K, x^{(i)}_H) + b_H).
\end{equation}
Since the activations $\phi_H$, weights $W_H$ and biases $b_H$ are kept in the host and the latent variables are not shared to the guests, no guest can infer the data of host based on Eq. (\ref{host_forward}) without knowing the parameters.
For the host, the latent variables are encoded in the guest by
\begin{equation}\label{guest_forward}
    a^{(i)}_k = \phi_k (W_k (x^{(i)}_k) + b_k).
\end{equation}
Knowing the latent variables $a^{(i)}_k$ only cannot infer the data $x^{(i)}_k$ without other unshared activations $\phi_k$, weights $W_k$ and biases $b_k$ from Eq. (\ref{guest_forward}). Unlike other FL algorithms, the encoder of guests is pre-trained via self-taught learning and parameters update are not required during the joint training. This indicates that the gradients of the respective neural network, $\frac{\partial L}{\partial W_H}$ and $\frac{\partial L}{\partial a_k}$ , are not known by the others. This prevents potential information leakage as the parties may infer data by stacking the gradients \cite{campbell2009generalized}. Hence, the parameters of activations, weights and biases act as the key of the latent variables and the proposed algorithm is computationally secured without sharing this high dimensional key.

\subsubsection{Statistical Security}
We purposely use VAE as the feature extractor for statistical security. Since the VAE is trained to regulate the distribution of latent space to the Gaussian distribution by minimizing their Kullback-Leibler (KL) divergence, we can expect that the distribution of latent variables is similar to Gaussian. From the perspective of the host, the received latent variables are Gaussian distribution. In other words, the statistical distance between the distribution of latent space and Gaussian distribution is small. A scheme is said to be $\varepsilon$-statistically secure if statistical distance between two encrypted plaintexts is smaller than $\varepsilon$ \cite{crepeau2008statistical}. Therefore, the algorithm is statistical secure using a well-trained VAE and transmitting the latent variables from the guests to the host will not reveal the raw data.

\section{Experiments}
In this section, we evaluate the empirical performance of STFL algorithm. First, we introduce the experimental settings and the datasets for the classification tasks. Second, we demonstrate the performance of STFL algorithm compared to other state-of-the-art methods.

\subsection{Experimental Settings and Datasets}

The STFL algorithm was built on an industrial-grade federated learning framework called Federated AI Technology Enabler (FATE) \cite{FATE}. Since our purpose is to demonstrate STFL's ability on vertical federated learning tasks with heterogeneous datasets, rather than challenging the upper limits of classification accuracy, we avoid using complex deep neural network structures and use a simple multilayer perceptron as the main structure in the experiments. For the self-taught learning component of the guest, a multilayer perceptron with $n_g$ hidden nodes is used to train as a VAE, where $n_g$ is equal to $5$ times its number of input features. The length of latent vector $n_z$ is limited to half of its input features. After self-taught training, the decoder will be removed and the latent vector $z$ will be inputted to host's network together with the input data of the host. Host's network contains $n_h$ hidden nodes, where $n_h$ is equal to $5$ times its number of input features. The following methods and neural network structure are compared with STFL:
\begin{itemize}
    \item Centralized: A multilayer perceptron with $n_h$ hidden nodes was used for the classification tasks, where $n_h$ is equal to $5$ times its number of input features. The data were combined and the multilayer perceptron was trained in a centralized manner as a baseline. 
    \item Hierarchical: Multilayer perceptrons with the same structure of STFL were used in this method. The main difference between this method and STFL is that the guest's network is not self-trained. The network was jointly trained with the host's network by the combined data in a centralized manner. The parameters of guest's network can be updated by the gradients obtained during joint training.
    \item FATE: The heterogeneous neural network module in FATE was used. It uses homomorphic encryption to encrypt the transmitted values so that the gradient can back-propagate to the network without revealing the data privacy. The algorithm is built in FATE based on \cite{zhang2018gelu} and \cite{zhang2019additively}. It contains bottom layers with $n_g$ hidden nodes for each party, an interactive layer with $n_h$ hidden nodes, and a top model with $n_h$ hidden nodes.
\end{itemize}
The batch size is 128. We use Adam \cite{kingma2015adam} as the optimizer with $0.001$ learning rate. The training lasts for 100 epochs. The binary cross-entropy is used as the loss function.

The following datasets for classification task provided in FATE were used in the experiments:
\begin{itemize}
    \item Breast Cancer Wisconsin Diagnostic Dataset (cancer)\footnote{\url{https://www.kaggle.com/uciml/breast-cancer-wisconsin-data}}: The goal of this task is to determine whether the cancer is benign or malignant based on the features captured from the cell nucleus, including radius, texture, perimeter, area, smoothness, compactness, concavity, concave points, symmetry, and fractal dimension. The mean, standard error, and largest of the features are presented in the dataset, which results in 30 features for each data. It contains 569 data samples.
    \item Default Payment of Credit Card Clients Dataset (payment)\footnote{\url{https://archive.ics.uci.edu/ml/datasets/default+of+credit+card+clients}}: The goal of this dataset is to determine whether the clients are credible or not based on 23 features, including the amount of given credit, gender, education, marital status, age, history of past payments, amount of bill statements, and amount of previous payments. It contains 30,000 data samples.
    \item Give Me Some Credit Dataset (credit)\footnote{\url{https://www.kaggle.com/c/GiveMeSomeCredit/data}}: The goal of this task is to predict the probability that somebody will experience financial distress in the next two years to improve the credit scoring system. The dataset contains 10 features and a binary label indicating whether the person faces financial distress in the next two years. It contains 150,000 data samples.
\end{itemize}
The features are vertically divided into datasets for the host and guest as provided in FATE. To show the ability of our proposed STFL which supports self-train in advance to the joint training, we use the $40\%$ data in the dataset as the self-taught set for the guests and the corresponding $40\%$ data in host are disregarded since the ID are the same. In practice, the guest can collect public data or use the data with different ID as the self-taught set. The $40\%$ self-taught set is randomly selected from the dataset and other $40\%$ data is randomly selected as the training set. The remaining $20\%$ data is used as the testing set.

\begin{table}[!t]
\renewcommand{\arraystretch}{1.3}
\centering
\caption{Classification Accuracy}\label{accuracy_table}
\begin{tabular}{l|c|c|c}
\hline\hline
 & cancer & payment & credit \\
\hline
Centralized & 91.15\% & 81.10\% & 93.16\% \\
\hline
Centralized (all data) & 92.92\% & 81.32\% & 93.39\% \\
\hline
Hierarchical & 92.92\% & 78.20\% & 93.54\% \\
\hline
Hierarchical (all data) & 92.92\% & 78.20\% & 93.54\% \\
\hline
FATE & 92.04\% & 74.87\% & 93.44\% \\
\hline
FATE (all data) & 95.58\% & 74.85\% & 93.24\% \\
\hline
STFL & 92.92\% & 78.50\% & 93.41\% \\
\hline\hline
\end{tabular}
\end{table}

\begin{table}[!t]
\renewcommand{\arraystretch}{1.3}
\centering
\caption{Confusion Matrix}\label{confusion_matix}
\subfloat[FATE with ``cancer'' dataset]{
\begin{tabular}{c|c|c|c}
\hline\hline
\multicolumn{2}{c|}{} & \multicolumn{2}{c}{Predicted class} \\
\cline{3-4}
\multicolumn{2}{c|}{} & Class 1 & Class 2 \\
\hline
\multirow{2}{*}{True class} & Class 1 & 37 & 6 \\
\cline{2-4}
& Class 2 & 3 & 67 \\
\hline\hline
\end{tabular}
\label{FATE_breast}}
\hspace{\fill}
\subfloat[STFL with ``cancer'' dataset]{
\begin{tabular}{c|c|c|c}
\hline\hline
\multicolumn{2}{c|}{} & \multicolumn{2}{c}{Predicted class} \\
\cline{3-4}
\multicolumn{2}{c|}{} & Class 1 & Class 2 \\
\hline
\multirow{2}{*}{True class} & Class 1 & 37 & 6 \\
\cline{2-4}
& Class 2 & 2 & 68 \\
\hline\hline
\end{tabular}
\label{STFL_breast}}
\hspace{\fill}
\subfloat[FATE with ``payment'' dataset]{
\begin{tabular}{c|c|c|c}
\hline\hline
\multicolumn{2}{c|}{} & \multicolumn{2}{c}{Predicted class} \\
\cline{3-4}
\multicolumn{2}{c|}{} & Class 1 & Class 2 \\
\hline
\multirow{2}{*}{True class} & Class 1 & 4197 & 494 \\
\cline{2-4}
& Class 2 & 1014 & 295 \\
\hline\hline
\end{tabular}
\label{FATE_default}}
\hspace{\fill}
\subfloat[STFL with ``payment'' dataset]{
\begin{tabular}{c|c|c|c}
\hline\hline
\multicolumn{2}{c|}{} & \multicolumn{2}{c}{Predicted class} \\
\cline{3-4}
\multicolumn{2}{c|}{} & Class 1 & Class 2 \\
\hline
\multirow{2}{*}{True class} & Class 1 & 4484 & 207 \\
\cline{2-4}
& Class 2 & 1083 & 226 \\
\hline\hline
\end{tabular}
\label{STFL_default}}
\hspace{\fill}
\subfloat[FATE with ``credit'' dataset]{
\begin{tabular}{c|c|c|c}
\hline\hline
\multicolumn{2}{c|}{} & \multicolumn{2}{c}{Predicted class} \\
\cline{3-4}
\multicolumn{2}{c|}{} & Class 1 & Class 2 \\
\hline
\multirow{2}{*}{True class} & Class 1 & 27739 & 251 \\
\cline{2-4}
& Class 2 & 1716 & 294 \\
\hline\hline
\end{tabular}
\label{FATE_give}}
\hspace{\fill}
\subfloat[STFL with ``credit'' dataset]{
\begin{tabular}{c|c|c|c}
\hline\hline
\multicolumn{2}{c|}{} & \multicolumn{2}{c}{Predicted class} \\
\cline{3-4}
\multicolumn{2}{c|}{} & Class 1 & Class 2 \\
\hline
\multirow{2}{*}{True class} & Class 1 & 27550 & 440 \\
\cline{2-4}
& Class 2 & 1538 & 472 \\
\hline\hline
\end{tabular}
\label{STFL_give}}
\end{table}

\subsection{Classification Accuracy}
Table \ref{accuracy_table} shows the testing accuracy with the tested datasets. For the methods marked with (all data), it means the $40\%$ self-taught set is incorporated into the training set to evaluate the accuracy with all available data.
For ``cancer'' dataset, FATE trained by all data has the highest accuracy while all other methods have a similar accuracy. From the experiments of using ``cancer'' dataset, we can see that STFL is robust to the data size since it contains very small (228) data samples for training the VAE as compared to other two large datasets.
For ``payment'' dataset, the centralized method trained with all data has the highest accuracy when compared to other methods. In general, most of the methods have similar accuracy except FATE has a lower accuracy. For ``credit'' dataset, the hierarchical method has the highest accuracy when compared to other methods. In general, the accuracy of all methods is similar. We can see that although STFL cannot achieve the highest accuracy, it is not the method with the lowest accuracy. It has similar accuracy with other state-of-the-art methods. Therefore, fixing the parameters in the guest's network during training without gradient back-propagation does not significantly decrease its prediction accuracy since it has been pre-trained.

Table \ref{confusion_matix} shows the confusion matrices of the six experiments of FATE and STFL methods with ``cancer'', ``payment'' and ``credit'' datasets. For ``cancer'' and ``payment'' dataset, both methods have a similar results as shown in Tables \ref{FATE_breast} to \ref{STFL_default}. For results using ``credit'' dataset as shown in Tables \ref{FATE_give} and \ref{STFL_give}, although the resulting accuracy are similar for FATE and STFL, STFL tends to predict more samples since Class 2 when compared to FATE. This may due to the fact that the classes are imbalanced in ``credit'' dataset and FATE poorly generalizes the model for this imbalanced dataset. Hence, STFL's accuracy still comparable to FATE without the back-propagation parameter update of guest's network.

\begin{table}[!t]
\renewcommand{\arraystretch}{1.3}
\centering
\caption{Training time}\label{time}
\begin{tabular}{c|c|c|c}
\hline\hline
 & cancer & payment & credit \\
 \hline
FATE & 1,737s & 137,957s & 191,109s \\
\hline
STFL & 88s & 941s & 7,723s \\
\hline\hline
\end{tabular}
\end{table}

\subsection{Training speed}
Table \ref{time} shows the required time for training the multilayer perceptron using different methods. For different datasets, the time is different since the data size of each dataset is different. The larger the dataset, the longer the required training time. FATE requires much more training time compared to STFL for the three datasets. The main reason is that FATE uses homomorphic encryption for the shared information, which is computationally expensive. On the other hand, STFL avoids using homomorphic encryption and use a self-trained encoder of VAE to transform features from the data to the latent space which can be computed in a much shorter time. Therefore, the computational time of STFL is much short than that of FATE.

\section{Conclusion}
FL can facilitate privacy-preserving data sharing among data owners to enhance AI modeling and applications. However, state-of-the-art FL technologies for heterogeneous datasets suffer from extremely low training speed which is impractical for real-world applications. Inspired by self-taught learning, we propose a new FL method for heterogeneous datasets called STFL to exploit unsupervised feature extraction techniques for the federated learning with heterogeneous datasets while preserving data privacy. Experiments on different datasets were preformed to evaluate the validity and efficiency of the proposed method. The results show that STFL dramatically reduces the required training speed while maintaining a similar level of prediction accuracy in the experiments.

Achieving the optimal balance between privacy and learning efficiency is still an open question to federated learning. In future, more research needs to be done on optimizing the learning efficiency while preserving data privacy.

\bibliographystyle{IEEEtran}
\bibliography{references}

\end{document}